\documentclass[10pt,conference]{IEEEtran}
\usepackage{cite}
\usepackage{amsmath,amssymb,amsfonts}       % blackboard math symbols
\usepackage{algorithmic}
\usepackage{graphicx}
\usepackage{textcomp}
\usepackage{xcolor}
\def\BibTeX{{\rm B\kern-.05em{\sc i\kern-.025em b}\kern-.08em
    T\kern-.1667em\lower.7ex\hbox{E}\kern-.125emX}}

\usepackage[utf8]{inputenc} % allow utf-8 input
\usepackage[T1]{fontenc}    % use 8-bit T1 fonts
\usepackage{hyperref}       % hyperlinks
\usepackage{url}            % simple URL typesetting
\usepackage{booktabs}       % professional-quality tables
\usepackage{nicefrac}       % compact symbols for 1/2, etc.
\usepackage{microtype}      % microtypography

\usepackage{enumitem}
\usepackage{multirow}

\usepackage{fancyvrb}
\RecustomVerbatimEnvironment{verbatim}{Verbatim}{fontsize=\small}

\usepackage[scaled=.8]{beramono}

\usepackage{listings}

\begin{document}

\title{NNStreamer: Efficient and Agile Development of On-Device AI Systems}

\author{\IEEEauthorblockN{MyungJoo~Ham\IEEEauthorrefmark{1},
{Jijoong~Moon},
{Geunsik~Lim},
{Jaeyun~Jung},
{Hyoungjoo~Ahn},
{Wook~Song},}
\IEEEauthorblockN{
{Sangjung~Woo},
{Parichay~Kapoor},
{Dongju~Chae},
{Gichan~Jang},
{Yongjoo~Ahn}, and 
{Jihoon~Lee}
}\\

\IEEEauthorblockA{Samsung Research, \\Samsung Electronics\\Seoul, Korea\\
\{myungjoo.ham, jijoong.moon, geunsik.lim, jy1210.jung, hello.ahn, wook16.song,\\
sangjung.woo, pk.kapoor, dongju.chae, gichan2.jang, yongjoo1.ahn, jhoon.it.lee\}@samsung.com
}
}

\maketitle

\begin{abstract}
We propose \textit{NNStreamer}, a software system that handles neural networks as filters of stream pipelines,
applying the stream processing paradigm to deep neural network applications.
A new trend with the wide-spread of deep neural network applications is on-device AI.
It is to process neural networks on mobile devices or edge/IoT devices instead of cloud servers.
Emerging privacy issues, data transmission costs, and operational costs signify the need for on-device AI, especially if we deploy a massive number of devices.
\textit{NNStreamer} efficiently handles neural networks with complex data stream pipelines on devices, significantly improving the overall performance with minimal efforts.
Besides, \textit{NNStreamer} simplifies implementations and allows reusing off-the-shelf media filters directly, which reduces developmental costs significantly.
We are already deploying \textit{NNStreamer} for a wide range of products and platforms, including the Galaxy series and various consumer electronic devices.
The experimental results suggest a reduction in developmental costs and enhanced performance of pipeline architectures and \textit{NNStreamer}.
It is an open-source project incubated by Linux Foundation AI, available to the public and applicable to various hardware and software platforms.
% @lim: nnstreamer 실험결과를 한 줄 정도는 작성하여 요약할 필요가 있을거 같습니다. 우리쪽 실험결과를 한줄로 abstract에서 자랑하는게 좋지 않을까? 싶습니다. 

%Many usage scenarios of on-device AI address real-time data streams from sensors of the device itself: e.g., cameras and microphones.
%For such data streams, users and governments do not want manufacturers to access the live feeds of the streams, manufactures and operators do not want to send and receive such huge traffic of streams incurring high latency, and they also do not want to spend excessively for transmissions and servers especially when they have extremely many devices deployed.

\end{abstract}

\begin{IEEEkeywords}
neural network, on-device AI, stream processing, pipe and filter architecture, open source software
\end{IEEEkeywords}

\renewcommand{\thefootnote}{\roman{footnote}}
\footnotetext[0]{\IEEEauthorrefmark{1}The corresponding author.}
\section{Introduction}\label{S_Intro}

We have witnessed the proliferation of deep neural networks in the last decade.
With the ever-growing computing power, embedded devices start to run neural networks, often assisted by hardware accelerators~\cite{APPLE_A13_NEWS_2020, chen2018tvm, Moreau:vta-tvm, ionica2015movidius, Jouppi:TPU, npu-isvlsi, ignatov2018ai, exynos9-9820}. 
Such accelerators are already common in the mobile industry~\cite{exynos9-9820,APPLE_A13_NEWS_2020}.
Running AI mechanisms directly on embedded devices is called on-device AI~\cite{MITTechReview}.
On-device AI can be highly attractive with the following advantages of in-place data processing.
\begin{itemize}
\item Avoid data privacy and protection issues by not sharing data with cloud servers.
\item Reduce data transmissions, which can be crucial for processing video streams in real-time.
\item Save operating costs of servers, especially crucial with millions of devices deployed.
\end{itemize}

Limited computing power, high data bandwidth, and short response time are significant challenges of on-device AI: e.g., 
AR Emoji~\cite{SamsungAREmoji}, Animoji~\cite{AppleAnimoji}, robotic vacuums, and live video processing.
With more sophisticated AI applications, multiple input streams and neural networks may exist simultaneously as in the complex camera systems of high-end smartphones of today.
Numerous neural networks may share inputs, and outputs of a neural network may be inputs of others or the network itself.
Composing a system with multiple networks allows training and reusing smaller networks, which may reduce costs, increase performance, enhance robustness, or help construct modular neural networks~\cite{schmidt1996modular,tahmasebi2011application,sanchez2015optimization}.
Managing data flows and networks may become highly complicated with interconnections of networks and other nodes along with fluctuating latency, complex topology, and synchronizations.
Such interconnections are data streams between nodes; thus, we may describe each node as a filter and a system as a pipeline, ``pipe and filter architectures''~\cite{ortega2005pipes}.

The primary design choice is to employ and adapt a multimedia stream processing framework for constructing neural network pipelines, not constructing a new stream framework.
The following significant problems and requirements, which are part of the observed ones of our on-device AI projects, have already been addressed by conventional multimedia frameworks for years:
\begin{enumerate}[label=P\arabic*.]
\item Fetching and pre-processing input streams may be extremely complicated; i.e., video inputs may have varying formats, sizes, color balances, frame rates, and sources determined at run-time.
Besides, with multiple data streams, processors, and algorithms, data rates and latency may fluctuate and synchronizing data streams may become extremely difficult.
\item Components should be highly portable. We have to reuse components and their pipelines for a wide range of products.
\item It should be easy to construct and modify pipelines even if there are filters executed in parallel requiring synchronization. The execution should be efficient for embedded devices. 
\item We want to reuse a wide range of off-the-shelf multimedia filters.
\end{enumerate}

%%% PK COMMENT 20201015
%%% P1-4 should somewhere also talk about fluctuating latency/synchronization. 2 paragraphs above, this is one of the main reasons of choosing "pipe and filter architectures". Also, it is solved by Gstreamer, so no issues on that end.
Some other significant problems and requirements are either introduced by reusing conventional multimedia stream processing frameworks (P5) or not addressed by such frameworks (P6 and P7).
\begin{enumerate}[label=P\arabic*.]
\setcounter{enumi}{4}
\item Input streams should support not only audio and video, but also general tensors and binaries. It should also support recurrences, which are not allowed by conventional media frameworks.
\item Different neural network frameworks (NNFW) such as TensorFlow and Caffe may coexist in prototypes. We want to integrate the whole system as a pipeline for such prototypes as well.
\item Easily import third-party components. Hardware accelerators often require converted neural networks with their dedicated format and libraries instead of general formats and NNFWs~\cite{COMPILER_cyphers2018intel,COMPILER_lin2019onnc}.
\end{enumerate}

We choose GStreamer~\cite{GStreamer} as the basis framework.
Gstreamer is a battle-proven multimedia framework for various products and services and has hundreds of off-the-shelf filters.
It is highly portable and modular, and virtually everything can be updated in a plug and play fashion.
To address P1 to P7, we provide numerous GStreamer plugins, data types, and tools, described in Section~\ref{S_Approach}, which allow interacting with various NNFWs, hardware accelerators, and other software components or manipulating stream paths and data.

%%%%%%%%%%%%%%%%%%%%%%%%%%%%%%%%%%%%%%%%%%%%%%%%%%
%%%%%%%%%%%%%%%%%%%%%%%%%%%%%%%%%%%%%%%%%%%%%%%%%%
%%%%%%%%%%%%%% REWRITE, REPHRASE THESE!!!!!!!!!!!!!!!!!!!!!
%%%%%%%%%%%%%%%%%%%%%%%%%%%%%%%%%%%%%%%%%%%%%%%%%%
%%%%%%%%%%%%%%%%%%%%%%%%%%%%%%%%%%%%%%%%%%%%%%%%%%
Our major contributions include:
\begin{itemize}
    \item Show that applying stream processing paradigm to complex multi-model and multi-modal AI systems is viable and beneficial in various environments, and provide an easy-to-use, efficient, portable, and ready-to-deploy solution.
    \item Provide standard representations of tensor data streams that interconnect different frameworks and platforms and off-the-shelf media filters to AI systems with minimal efforts, which efficiently allows processing complex pipelines and attaching AI mechanisms to applications.
    \item Allow developers to add arbitrary neural network frameworks, hardware accelerators, models, and other components easily with the given frameworks and code templates. Then, make the proposed mechanism product-ready and release it to various platforms and products. 
\end{itemize}

\textit{NNStreamer} is an open-source project incubated by Linux Foundation AI, released for Tizen, Android, Ubuntu, OpenEmbedded, and macOS.
\textit{NNStreamer} provides Machine Learning APIs of Tizen, an OS for a wide range of consumer electronics.
We are applying \textit{NNStreamer} to Android and Tizen products, as well.
It supports Tizen Studio natively (C and C\#) and Android Studio via JCenter, and Play Store offers \textit{NNStreamer} sample applications.

\section{Related work}\label{S_RWork}

\textbf{GStreamer}~\cite{GStreamer} is the multimedia framework of Tizen and many Linux distributions.
GStreamer provides APIs in various programming languages and utilities to construct, execute, and analyze media stream pipelines for different operating systems along with hundreds of off-the-shelf filters, which \textit{NNStreamer} inherits.
GStreamer is highly modular; every filter and path control is a plugin attachable in run-time.
Various systems, whose reliability and performance are crucial, use GStreamer.
For example, the BBC uses GStreamer for its broadcasting systems~\cite{BBCGSTCONF}.
Samsung (Tizen) and LG (WebOS) use it as the media engine of televisions.
Centricular uses it for TVs, set-top boxes, medical devices, in-vehicle infotainment, and on-demand streaming solutions~\cite{centricular-gstplayer,centricular-sync-multi-room,centricular-new-gststream-api,centricular-opengl-pipeline,centricular-zerocopy}.

\textbf{FFmpeg}~\cite{FFMPEG}, another popular multimedia framework, is not modular, and everything is built-in; thus, it is not suitable for our purposes.
\textbf{StageFright}~\cite{StageFright} is the multimedia framework of Android, depending on Android services.
Unlike GStreamer, it is not portable for general Linux systems and does not allow applications to construct arbitrary pipelines.
\textbf{AVFoundation}~\cite{AVFoundation} is the multimedia framework of iOS and macOS.
AVFoundation may provide input frames to Core ML~\cite{CoreML_API}, the machine learning framework of iOS and macOS, to construct a neural network pipeline.
However, app developers cannot apply neural networks as native filters of multimedia pipelines, and they need to implement interconnections between neural networks and multimedia pipelines.
\textbf{DirectShow}~\cite{directshow-new-media-ach} is the multimedia framework of Windows.
DirectShow and AVFoundation are proprietary software for proprietary platforms; thus, we cannot alter them for the given purposes.

Google has proposed \textbf{MediaPipe}~\cite{MEDIAPIPE_lugaresi2019mediapipe} to process neural networks as filters of pipelines.
It supports Linux, Android, and iOS, but it is not portable enough.
Its dependency on Google's in-house tool, Bazel, and inflexible library requirements make it not portable for embedded systems; i.e., it is hard to share system libraries with other software.
MediaPipe re-implements a pipeline framework and cannot reuse conventional media filters; thus, P1 to P4 are only partially met while P5 is not an issue.
Initially, it has targeted server-side AI services, not embedded devices; thus, P1, P2, and P4 might have been not considered.
Specifically, for in-house servers, they may restrict input formats (P1 and P4 are irrelevant) and consider homogeneous platforms and architectures (P2 is irrelevant).
Another issue is that MediaPipe allows only a specific version of TensorFlow as NNFWs; e.g., TensorFlow 2.1 for MediaPipe 0.7.4.
Such inflexibility makes integrating other NNFWs or hardware accelerators unnecessarily tricky.
In Section~\ref{S_Evaluation}, we show an example (E4) of how critical this can be.
We expect that they probably have more features hidden in-house; they have partially opened MediaPipe since 2019.
On the other hand, \textit{NNStreamer} has been fully opened since 2018.

Nvidia \textbf{DeepStream}~\cite{GSTCONF18_DEEPSTREAM, nvidia-deepstream} provides GStreamer plugins to process neural networks with NVidia's proprietary hardware; thus, P2 and P7 cannot be achieved while P1, P3, and P4 are achieved.
DeepStream addresses P5 indirectly and partially by embedding tensors in metadata of streams (no recurrence support).
In other words, DeepStream requires conventional media (audio/video/text) for inputs and does not consider tensors as first-class citizens of stream data.
Therefore, if the topology is complicated or inputs are arbitrary binaries, writing a pipeline can be difficult.

\textit{NNStreamer} provide interconnections between pipelines of different frameworks or remote nodes by proposing a standard tensor stream protocol via Flatbuf~\cite{FLATBUF} and Protobuf~\cite{PROTOBUF}.
\textit{NNStreamer} can collaborate with MediaPipe pipelines by embedding MediaPipe pipelines into \textit{NNStreamer} pipelines.

\section{Design and implementations}\label{S_Approach}

\begin{figure*}
\centering
\includegraphics[width=0.98\textwidth]{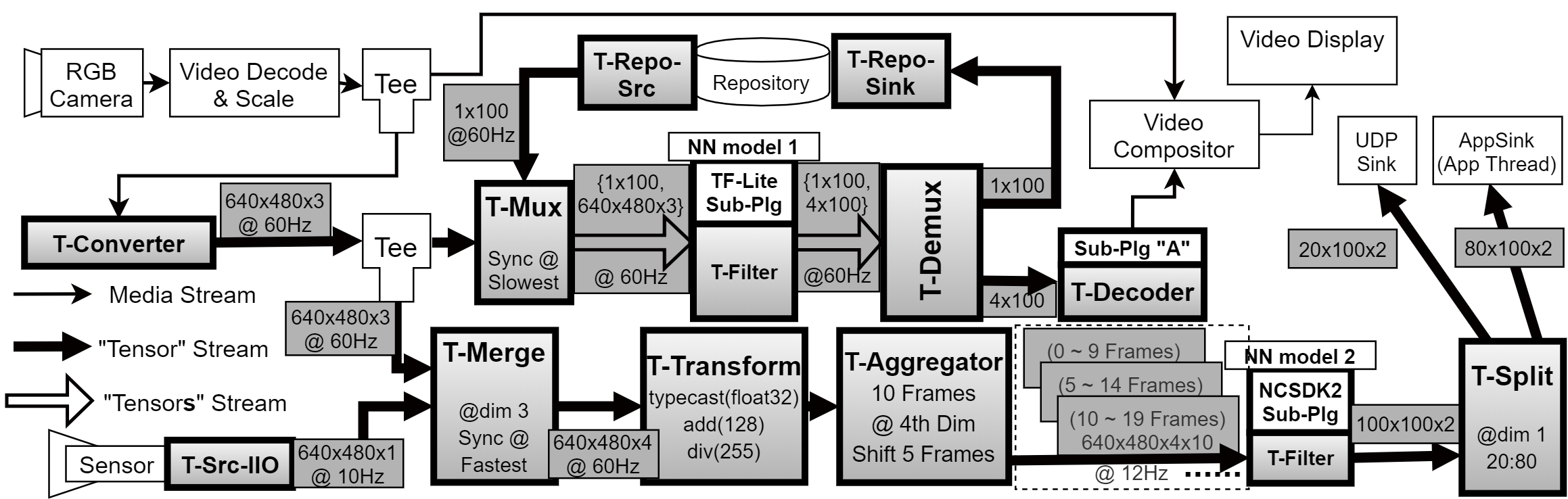}
\caption{An exemplary pipeline with representative \textit{NNStreamer} components.}
\label{FIG_EXPPLN_MajorFilters}
\end{figure*}

Each neural network model is an atomic filter of a pipeline (pipe and filter architecture~\cite{ortega2005pipes}).
We delegate executions of neural network models to their corresponding NNFWs, such as TensorFlow.
Delegation allows \textit{NNStreamer} to execute each model efficiently with P6 and P7 satisfied and to focus on how to describe and integrate interconnections and filters.
For example, to accelerate a TensorFlow model with GPU in a pipeline, users simply need to make sure that a compatible TensorFlow-GPU exists.
Similarly, installing libraries properly and writing a pipeline consisting of Vivante models allows accelerating the pipeline with a Vivante NPU~\cite{VIVANTE}.
This approach keeps the performance of and compatibility with off-the-shelf execution environments.

%%%%%%%%% NOT VERY RELEVANT
%If a whole system (such as an autonomous vehicle) is implemented as a single neural network, we may entirely rely on an NNFW.
%However, it is much more challenging to train, debug, and verify such a big end-to-end model than to do so with regular models.
%Moreover, combining regular models in a pipeline makes it much easier to update and debug the system.

We recognize tensors as first-class citizens of stream data, unlike DeepStream~\cite{nvidia-deepstream}, not limiting streams to conventional media, and add stream path controls for tensors.
We define two GStreamer data types: \texttt{other/tensor} and \texttt{other/tensor}\textbf{s}.
An \texttt{other/tensor} has an element type, dimensions, and a frame rate (e.g., \texttt{uint8}, \texttt{640:480:3}, \texttt{20 Hz}).
An \texttt{other/tensors} combines up to 16 (default limit of memory chunks in a frame) different tensors with a synchronized frame rate.
We store each tensor in an individual memory chunk so that mux and de-mux do not incur memory copies.

We do not express rank numbers in tensor stream types; thus stream types of compatible data formats (e.g., \texttt{640:480 [rank 2]} and \texttt{640:480:1:1 [rank 4]}) are considered to be equivalent by stream type checkers of GStreamer, which ``negotiates'' stream types between elements of pipelines in run-time.
However, there are a few NNFWs (e.g., TensorRT), which require to identify the rank numbers of input and output data as well as their dimensions and types.
Users may explicitly express rank numbers (e.g., ``\texttt{input=640:480}'' denotes rank 2 and  ``\texttt{input=640:480:1:1}'' denotes rank 4) in such cases to satisfy such NNFWs.

Figure~\ref{FIG_EXPPLN_MajorFilters} shows an exemplary pipeline demonstrating \textit{NNStreamer} components omitting a few trivial filters, such as queues.
Lightly shaded boxes with bold borders are \textit{NNStreamer} components.
Clear boxes are off-the-shelf components.
Shaded boxes with thin borders show properties of tensors.
In the figure, names are abbreviated; i.e., ``T'' denotes Tensor, the prefix of \textit{NNStreamer} filters.

The two neural networks in the figure, NN models, 1 and 2, use TensorFlow-lite and NCSDK2 sub-plugins (plugins of a plugin) of \textbf{Tensor-Filter} plugins, respectively.
\textit{NNStreamer} 1.6.0 of October 2020 provides sub-plugins for ARMNN, Caffe2, NNFW(ONE)-Runtime, OpenVINO, PyTorch, Qualcomm SNPE, Samsung SNAP, TensorFlow, TensorFlow-Lite, TensorRT, EdgeTPU, NCSDK2 (Movidius-X), Vivante, MediaPipe, and custom functions in C, C++, and Python.
Users may use such sub-plugins or write their own with the provided code templates and generators.

Tensor-Filter and its sub-plugin structure allow developers to use neural network models of the above frameworks with a unified interface even without pipelines as well.
In order to allow developers using the unified interface without pipelines, we provide ``Single API sets'' for Tizen (C/.NET) and Android (Java) products.

Inputs and outputs of Tensor-Filter are tensor streams.
\textbf{Tensor-Converter} converts media streams to tensor streams.
Sub-plugins of Tensor-Converters may accept unconventional (neither audio, video, nor text) data streams, e.g., Flatbuf~\cite{FLATBUF} or Protobuf~\cite{PROTOBUF} streams.
\textbf{Tensor-Decoder} may convert tensor streams to media or other data streams with sub-plugins, e.g., create a video stream of transparent backgrounds with boxes of detected objects or a Flatbuf stream from tensors.

\textbf{Tensor-Mux} bundles multiple \texttt{other/tensor} streams to an \texttt{other/tensors} stream.
\textbf{Tensor-Demux} un-bundles such a stream back to individual tensor streams.
\textbf{Tensor-Merge} creates an \texttt{other/tensor} (no ``s'') from multiple \texttt{other/tensor} streams, modifying dimensions.
\textbf{Tensor-Split} splits an \texttt{other/tensor} stream into multiple \texttt{other/tensor} streams.
From two 3x4 streams, Tensor-Merge creates a 6x4, 3x8, or 3x4x2 stream, and Tensor-Mux creates a \{3x4, 3x4\} stream.
Users may choose synchronization policies for Mux and Merge:
slowest (drop frames of faster sources), fastest (duplicate frames of slower sources), and base (keep the frame rate of the designated source).
\textbf{Tensor-Aggregator} merges frames temporally (e.g., merging frames $2i$ and $2i+1$, halving the frame rate), which may help implement LSTM or Seq2seq~\cite{sutskever2014sequence}.
All merging filters choose the latest timestamp.
A \textbf{Tensor-Repo-Src/Sink} pair may share a named repository to construct a recurring data path without a GStreamer stream, which prohibits cycles.
\textbf{Tensor-Src-IIO} creates tensor streams from Linux Industrial I/O~\cite{IIOWebPage} sensors.
\textbf{Tensor-Transform} applies operators to tensors: typecast, add/sub/mult/div, normalization, transpose, and so on.
Streams may be connected to and from application threads, networks, or files.
There are components not shown in the figure, as well.
\textbf{Tensor-ROS-Src}/\textbf{Sink} interact with ROS~\cite{quigley2009ros}, a popular robotics framework.
\textbf{Tensor-Src-TizenSensor} connects with Tizen Sensor Framework.
\textbf{Amcsrc} connects with Android MediaCodec (AMC).

Product engineers have added more technical requirements during commercialization.
\begin{itemize}
    \item Transparent and easy-to-apply parallelism, which is met and shown with experiments (E1).
    \item Dynamic pipeline topology, which is achieved by the nature of GStreamer.
    \item Interact with other frameworks such as ROS~\cite{quigley2009ros}, Android media framework, Tizen sensor framework, and Linux IIO~\cite{IIOWebPage}, which are achieved by \textit{NNStreamer} components.
    \item Dynamic flow control, which is mostly achieved by off-the-shelf filters if application threads may control flows directly: valve and input/output-selector.
    With \textbf{Tensor-If}, developers can control flows based on tensor values without the interventions of application threads.
    \item Rate override and QoS control, which is addressed by \textbf{Tensor-Rate}.
\end{itemize}

\section{Evaluations}\label{S_Evaluation}
We have released \textit{NNStreamer} for multiple consumer electronics prototypes and products of different software and hardware platforms.
Quality control teams have tested \textit{NNStreamer}, and related products will soon be available for consumers.
We show the following sets of experimental results:
\begin{enumerate}[label=E\arabic*.]
    \item Multi-model pipelines with AMLogic A311D SoC: 4 Cortex-A73 and 2 Cortex-A53 cores, 4 GiB RAM, and a Vivante neural processing unit (NPU, a hardware accelerator for neural networks).
    E1 demonstrates how efficiently and easily \textit{NNStreamer} utilizes different computing resources.
    E1 has the same configuration with some 2021 consumer electronics models.
    \item Activity Recognition Sensor (ARS) with Nexell S5P4418 SoC: 4 Cortex-A9 cores and 1 GiB RAM.
    E2 shows how easily and efficiently developers can implement and execute multi-modal and multi-model applications.
    ARS is deployed to hospitals and elderly care facilities.
    \item Multi-Task Cascaded Convolutional Networks (MTCNN) \cite{mtcnn-dai2016instance,mtcnn-zhang2016joint}.
    E3 evaluates an extremely complicated pipeline in various hardware platforms.
    \item Performance comparison against MediaPipe in a desktop PC (Intel i7-7700 and 16 GiB RAM).
\end{enumerate}

\begin{table*}
    \caption{E1 results of 100-second executions. I3 is Inception-v3, and Y3 is Yolo-v3. C/I3 uses CPU; others use NPU.\\Negative values show resource sharing overheads.}
    \label{tab:A311DComparisons}
    \centering
    \begin{tabular}{c l c c c c }
        Number of & \multirow{2}{*}{Configuration} & Throughput & CPU usage & Memory usage & Improved \\
        models & & (frames/s) & (\%) & (MiB) & throughput \\
        \hline
        \multirow{5}{*}{1} & a.Control / I3 & 19.4 & 161.8 & 84.5 & -- \\
         & b.Control / Y3 & 9.5 & 145.2 & 87.4 & --\\
         & c.NNStreamer / I3 & 28.0 & 17.0 & 24.5 & 44.3\% / a\\
         & d.NNStreamer / Y3 & 10.8 & 40.7 & 27.4 & 13.7\% / b\\
         & e.NNStreamer / C/I3 & 1.2 & 115.0 & 47.9 & --\\
        \hline
         \multirow{3}{*}{2} & f.NNStreamer / I3 + Y3 & 11.0, 7.0 & 44.7 & 32.6 & 4.5\% / c+d\\
         & g.NNStreamer / I3 + C/I3 & 27.8, 1.2 & 122.0 & 58.8 & \textcolor{red}{$-0.8$\%} / c+e \\
         & h.NNStreamer / Y3 + C/I3 & 10.5, 1.1 & 146.6 & 63.3 & \textcolor{red}{$-4.0$\%} / d+e \\
        \hline
         3 & i.NNS / I3 + Y3 + C/I3 & 11.0, 6.7, 1.1 & 151.7 & 68.4 & \textcolor{red}{$-2.3$\%} / c+d+e \\
    \end{tabular}
\end{table*}

\begin{figure*}
\centering\includegraphics[width=0.99\textwidth]{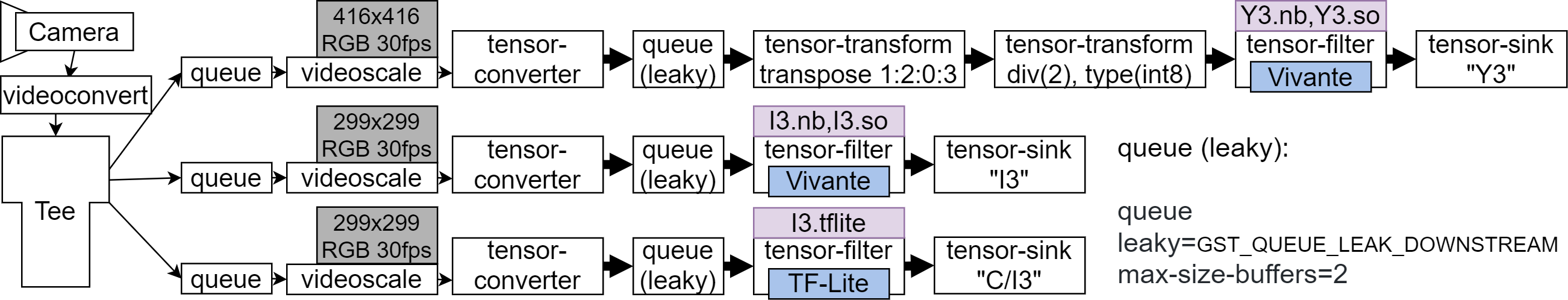}
\caption{Pipeline of E1 (case i in Table~\ref{tab:A311DComparisons}). Others (cases c to h) are sub-pipelines of i.}
\label{fig:E1_arch}
\end{figure*}

%%%%%%%%%%%%%%%%%%%%%%%% E1 %%%%%%%%%%%%%%%%%%%%%

\textbf{E1} in Figure~\ref{fig:E1_arch} evaluates the performance with heterogeneous resources: CPU and NPU.
Table~\ref{tab:A311DComparisons} compares the conventional implementation (Control) and various configurations of \textit{NNStreamer} pipelines (NNS) with 3000 input frames at 30 frames/s (fps).
Before the introduction of \textit{NNStreamer}, product engineers have implemented conventional code (Control), which processes every required operation serially for each input frame.
Higher performance (throughput), lower overheads (CPU, memory), and ease of implementation have successfully helped them abandon the conventional and adopt \textit{NNStreamer}.

Case c and d show performance improvement with the stream pipeline architecture.
Case e shows the base performance with CPU cores by using TensorFlow-lite instead of Vivante-NPU and its run-time libraries.
Case f shows that \textit{NNStreamer} can efficiently execute multiple models sharing an NPU with virtually no overheads; it has improved the overall performance by 4.5\%.
The output rates of individual models in g and h are virtually not affected while both are simultaneously executed; they suffer only 0.8\% and 4.0\% of overheads.
This result shows that \textit{NNStreamer} efficiently utilizes heterogeneous computing resources.
Case i shows that \textit{NNStreamer} can utilize shared resources and heterogeneous resources simultaneously and efficiently.
The improved throughput (or overhead of multi-model executions) is calculated by $ (\textrm{fps}(\textrm{I3})/\textrm{fps@c} + \textrm{fps}(\textrm{Y3})/\textrm{fps@d} + \textrm{fps}(\textrm{CPU-I3})/\textrm{fps@e}) / \textrm{\#HW} $, where $\textrm{\#HW}$ is 1 in f and 2 in g to i.
The capability to execute multiple models in parallel with virtually no overheads and with higher performance combined with the ease and flexibility of writing pipeline applications has been the reason for replacing the conventional implementations with \textit{NNStreamer}.
The memory usage cannot be compared against a and b directly because a and b are too inefficient, caching everything in memory.
Comparing h to \{f, g, h\} or \{f, g, h\} to \{c, d, e\} suggests that a larger pipeline with multiple models may be much more efficient than individual pipelines with a single model, especially if they are executed serially.

%%%%%%%%%%%%%%%%%%%%%%%% E2 %%%%%%%%%%%%%%%%%%%%%

\begin{figure*}
\centering\includegraphics[width=0.95\textwidth]{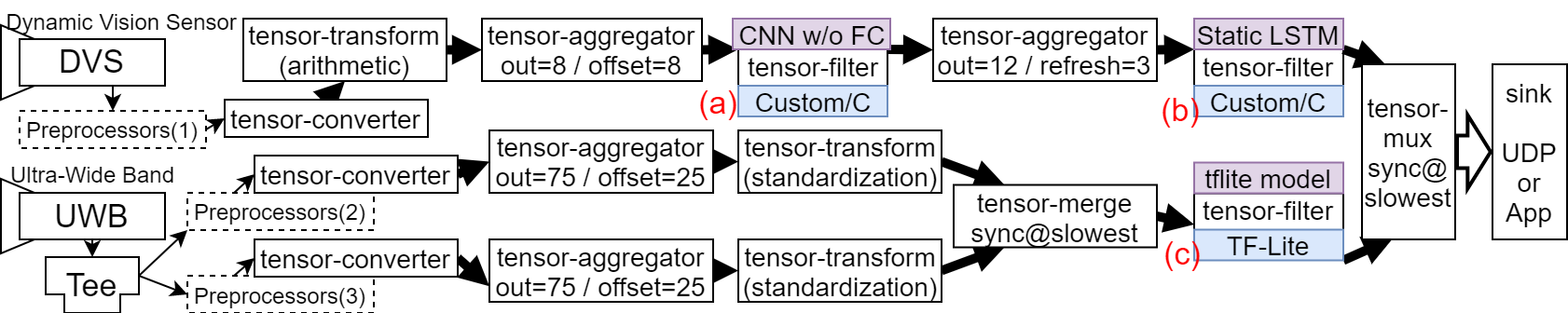}
\caption{Pipeline of E2, which is a multi-modal and multi-model pipeline for ARS devices.}
\label{FIG_EVAL_LSTM_Sensors}
\end{figure*}

%%%%%%%%%%%%%% RECONFIRM PIPELINE OF ARS. Duplicated sub-pipelines for tf-lite pass.
%%%%%%%%%%%%%% Then, redraw (rename a few elements!)

\textbf{E2} evaluates the performance and developmental efficiency of ARS, whose pipeline is shown in Figure~\ref{FIG_EVAL_LSTM_Sensors}, which consists of multiple sensors and neural networks.
Input stream feeding and synchronizations of such pipelines are not trivial and have a significant impact on performance and reliability.
The introduction of \textit{NNStreamer} to ARS has significantly reduced developmental effort.
Before \textit{NNStreamer}, a few developers have been implementing the pipeline partially for several weeks.
Then, with \textit{NNStreamer}, one developer has completed the pipeline within a few hours (only a dozen lines of codes) and optimized its performance by tweaking parameters within a couple of days.

The \textit{NNStreamer} pipeline runs faster and more efficiently, as well.
The memory usage is reduced by 48\% (448 MiB to 234 MiB).
The CPU workload with 30 fps live inputs is reduced by 43\% (90.43\% to 51.35\%), and both do not have frame drops.
The batch processing rate for recorded inputs is improved by 65.5\% with \textit{NNStreamer}: 46.0 to 59.4 in (a), 2.5 to 3.2 in (b), and 9.3 to 25.5 in (c) of Figure~\ref{FIG_EVAL_LSTM_Sensors}.
Note that because of aggregators, (b) and (c) process at slower rate.

%%%%%%%%%%%%%%%%%%%%%%%% E3 %%%%%%%%%%%%%%%%%%%%%

\begin{figure*}
\centering\includegraphics[width=0.99\textwidth]{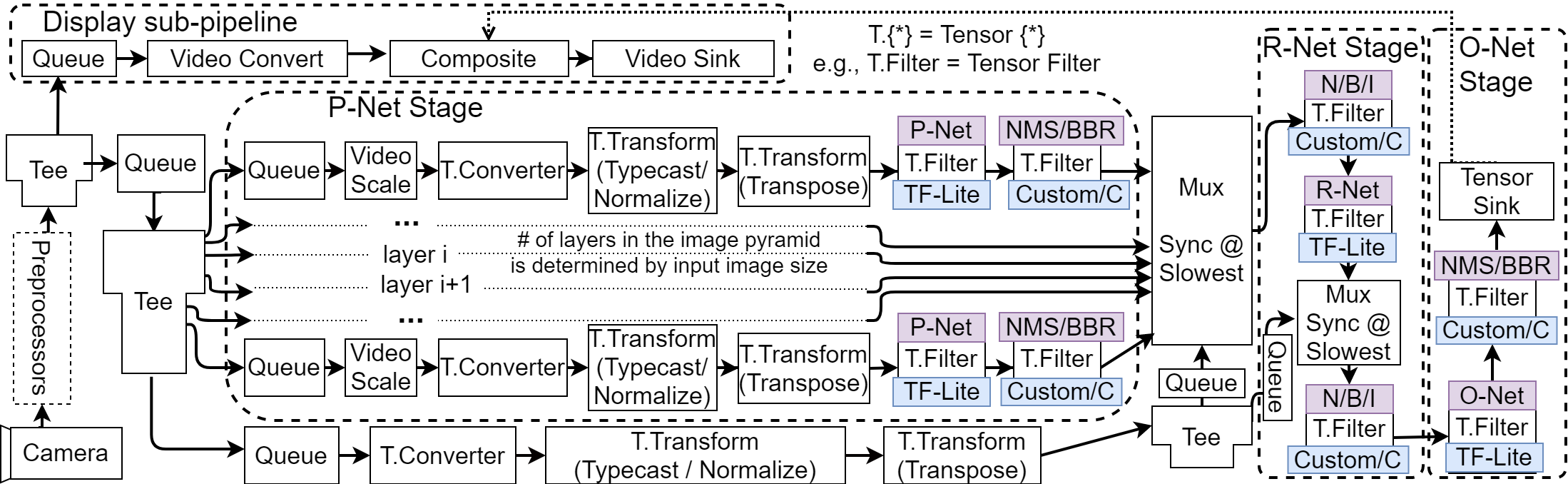}
\caption{Pipeline of E3 (MTCNN). N denotes non-maximum suppression (NMS). B denotes bounding box regression (BBR). I denotes image patch generation.}
\label{FIG_OVERVIEW_MTCNN}
\end{figure*}

\textbf{E3} evaluates MTCNN performance.
Figure~\ref{FIG_OVERVIEW_MTCNN} describes the complex topology of MTCNN.
The output is a video display (``Video Sink'') showing both camera inputs and inference results simultaneously.
We compare C/C++ implementations of \textit{NNStreamer} and ROS~\cite{quigley2009ros} (Control) with Full-HD videos in various devices: \texttt{A} (mid-end embedded, Exynos 5422), \texttt{B} (high-end automotive embedded, Exynos 8890), and \texttt{C} (PC with Intel i7-7700).
A ROS-fluent team (independent from the \textit{NNStreamer} team) is assigned for Control so that developers would implement efficient codes for Control.

There are several sub-pipelines with neural networks and merging points, which require synchronization and stream throttling.
For example, in P-Net Stage, processing a layer much faster (e.g., 30 fps) than other layers (e.g., 15 fps) is meaningless and deteriorates the overall performance.
Exploiting parallelism with proper synchronization and throttling becomes trivial with \textit{NNStreamer} as in E2: e.g., a dozen lines of C codes describe P-Net Stage.

\begin{table*}
\caption{MTCNN (E3) performance. Improvement is based on the geometric mean of ratios.}
\label{TBL_RESULT_MTCNN_PERF}
\begin{center}
\begin{tabular}{l r r r r r r r}
     {} &
     \multicolumn{2}{c}{\texttt{A} / Mid-end} &
     \multicolumn{2}{c}{\texttt{B} / High-end} &
     \multicolumn{2}{c}{\texttt{C} / PC} &
     Improved by\\
     {} & 
     Control & \textit{NNStreamer} &
     Control & \textit{NNStreamer} &
     Control & \textit{NNStreamer} &
     \textit{NNStreamer} (\%) \\
         &&&&\vspace{-1em}\\
         \hline
         &&&&\vspace{-1em}\\ 
    1. Throughput (fps) & 1.01 & 1.73 & 1.48 & 4.02 & 10.41 & 13.76 & 82.21 \\
    2. Overall latency (ms) & 981.8 & 811.0 & 704.5 & 539.4 & 94.3 & 85.9 & 16.79 \\
    3. P-Net latency (ms) & 795.7 & 531.5 & 614.3 & 358.1 & 74.3 & 41.1 & 40.06 \\
    4. R-Net latency (ms) & 82.4 & 174.4 & 67.7 & 101.4 & 9.7 & 25.9 & \textcolor{red}{$-6.61$} \\
    5. O-Net latency (ms) & 103.6 & 105.2 & 91.1 & 80.2 & 10.3 & 18.9 & \textcolor{red}{$-18.08$} \\
\end{tabular}
\end{center}
\end{table*}

Table~\ref{TBL_RESULT_MTCNN_PERF} compares the performance of \textit{NNStreamer} and Control.
\textit{NNStreamer} (\textit{NNS}) has lower overall latency (measured with 1 fps inputs) and higher throughput (measured with 30 fps inputs).
Row 2, compares the performance without the impact of pipeline data-parallelism~\cite{1990PipelinedDataPar} by processing a single input frame at a time, but with the effects of functional parallelism at P-Net (row 3) despite slower R-Net and O-Net (row 4 and 5).
The \textit{NNStreamer} case has 1959 lines of C codes (1004 of them are re-implementation of non-max suppression, bounding box regression, and image patch), which is slightly longer than Control (1644 lines of C++ codes).
Note that the \textit{NNStreamer} case supports exception handling and dynamic layers and video formats, which are not supported by Control, and provides a higher degree of parallelism.

%%%%%%%%%%%%%%%%%%%%% E4 %%%%%%%%%%%%%%%%%%%%%%%%%%

% preliminary results.... (HJ will update with better)
\begin{figure*}
\centering\includegraphics[width=0.99\textwidth]{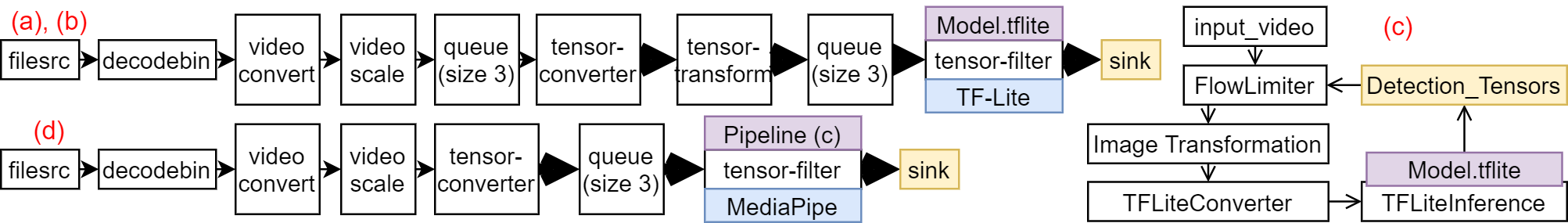}
\caption{Pipelines of E4. \textit{NNStreamer} with 2 TensorFlow versions: a and b, MediaPipe: c, hybrid: d.}
\label{FIG_E4_ARCH}
\end{figure*}

\begin{table*}
    \caption{E4 results with 10 executions of 1818 full-HD frames. $\pm$ shows standard deviations.}
    \label{tab:EVAL_E4}
    \centering
    \begin{tabular}{l r r r r}
         \multicolumn{2}{r}{(a) NNStreamer-a} &
           \multicolumn{1}{c}{(b) NNStreamer-b} &
           \multicolumn{1}{c}{(c) MediaPipe} &
           \multicolumn{1}{c}{(d) Hybrid}\\
%         & mean & \sigma & mean & \sigma & mean & \sigma & mean & \sigma \\
         \hline
      1. CPU (\%)   & 352.8 $\pm$ 0.44 & 168.7 $\pm$ 0.13 & 168.2 $\pm$ 0.08 & 168.0 $\pm$ 0.11 \\
%      2. VmRSS (kiB) & 204,354 & 204,005 & 187,716 & 301,913 \\
%      3. Mmap (B) & 129,471,106 & 126,540,822 & 1,312,817,824 & 776,418,434 \\
      %3. Mmap (MiB) & 139.5 & 136.7  & 1,216.7 & 740.5 \\ %%%%%%% NEED TO REWRITE NUMBERS
      2. Throughput (fps) & 46.9 $\pm$ 0.14 & 13.8 $\pm$ 0.03 & 13.3 $\pm$ 0.03 & 12.8 $\pm$ 0.01 \\
%      6. Latency (us) & 22,288 & 72,764 & 74,683 & 76,389 \\
      3. Latency (ms) & 20.8 $\pm$ 1.21 & 72.7 $\pm$ 2.05 & 74.5 $\pm$ 2.20 & 76.3 $\pm$ 2.18 \\
      4. Mem access (billions) & 21.9 $\pm$ 0.11 & 21.8 $\pm$ 0.05 & 23.5 $\pm$ 0.04 & 25.3 $\pm$ 0.08\\
      5. Mem size (MiB) & 199.5 $\pm$ 5.60 & 194.9 $\pm$ 0.32 & 185.1 $\pm$ 0.39 & 300.4 $\pm$ 3.36 \\
%       miss                ref(hit)            sum
%       sd                  sd
% a:    3,557,927,162       18,343,576,486       21,901,503,649 
%       60403395.53         67,742,482            109,562,447 
% b:     3,499,227,710      18,253,549,527           21,752,777,237 
%       11778148.17         47,730,616               51,556,365 
% c:     3,967,423,151 	 19,526,838,167              23,494,261,318 
%       21584923.41     35,276,972                   39,310,986 
% d:     4,214,481,030 	 21,062,937,302              25,277,418,332 
%       3921213.497	 79,096,564                  78,701,138 
% preproc/nns:       1,221,778,824 	 2,577,168,709       3,798,947,533 
%               2726156.05	 21,818,617                  21,768,163 
% preproc/mp:        1,782,935,549 	 3,449,153,906       5,232,089,455 
%               7251416.903	 43,103,943              43,623,937 

    \end{tabular}
\end{table*}

\textbf{E4} compares the performance of a MediaPipe sample pipeline and its equivalent \textit{NNStreamer} pipeline.
The neural network model for E4 is ``\texttt{ssdlite\_object\_detection.tflite}'', a MediaPipe reference model.
Figure~\ref{FIG_E4_ARCH} shows the 4 cases: a and b test \textit{NNStreamer} pipelines of different TensorFlow-lite versions (a: 1.15.2 and b: 2.1), c tests a MediaPipe pipeline, and d tests an \textit{NNStreamer} pipeline that has the pipeline c embedded as a filter.
We have removed some queues from c and d because they deteriorate their performance.
The cycle from ``Detection\_Tensors'' feeds the flow status so that FlowLimiter may throttle input rates.
\textit{NNStreamer} does not need it because GStreamer already has a bi-directional metadata stream channels for QoS controls embedded in the unidirectional data stream, which is why a stream path cycle is prohibited.

Table~\ref{tab:EVAL_E4} shows the benchmark results of E4.
Comparing (b) and (c) indicates that \textit{NNStreamer} has higher throughput (3.8\%) and lower latency (2.4\%).
More significantly, if \textit{NNStreamer} uses TensorFlow-lite 1.15.2 (a) instead of 2.1 (b), the throughput is more than tripled (x3.54), and the latency is almost quartered ($1 / 3.67$).
This improvement demonstrates the significant disadvantage of the inflexibility; i.e., MediaPipe of May 2020 (commit b6e680647) is strictly bound with TensorFlow 2.1 by its build system.
In other words, the ability to choose different NNFW versions may enhance the performance dramatically.
Therefore, the inflexibility, forfeiting P6, not only loses the compatibility with hardware accelerators and NNFWs but also misses the chances to perform better.

Another inefficiency comes from the primary design choice of re-implementing the pipeline framework.
This choice enforces to re-implement media filters and path controls; thus, abandoning opportunities for code reusing.
For 1818 input frames, if we execute pre-processors only, pre-processors of (b) and (c) consume CPU time (overhead) of 29.5 s and 41.4 s and real time (latency) of 9.86 s and 12.34 s, respectively; thus, MediaPipe's Open-CV re-implementations of media plugins perform 25\% worse and have 40\% of more overheads.
Inefficient pre-processing of (c) may be responsible for the performance deterioration, only partially; pipeline architectures can hide latency from throughput.
More critically, devices often have media processing hardware accelerators, which signifies the importance of off-the-shelf filters; i.e., mobile phones and TVs often have media decoders and format converters on chips and integrated with media frameworks.

In (d), pre-processors of both \textit{NNStreamer} and MediaPipe exist; however, those of MediaPipe have less workload (already pre-processed by \textit{NNStreamer}), which results in not-so-deteriorated performance even though both frameworks are simultaneously used.
This implies that \textit{NNStreamer} may import and execute arbitrary MediaPipe pipelines without a significant performance penalty.

%%%%%%%%%%%%%%%% NEED TO MENTION SOME NUMBERS HERE !!!!!!!!!!!
%  (excluding the effects of inefficient pre-processors--37.7\% more access--, by 1.7\%)
Row 4, the number of memory access measured by perf~\cite{weaver2013linux}, shows that MediaPipe accesses memory more by 8.0\%.
Massive memory accesses can significantly deteriorate the performance of embedded devices with NPUs, where NPUs enhance computing power, but memory bandwidth is limited; i.e., in E3, memory read bandwidths of a PC (\texttt{c}) and an embedded device (\texttt{a}) are measured to be 18.5 GiB/s and 2.6 GiB/s, respectively.
Row 5, memory size measured by peak VmRSS, suggests that \textit{NNStreamer} may consume a little more memory.
Memory size is affected by queues we have added to promote higher parallelism; each of the two queues may consume up to 17.8 MiB in (a) and (b).

%%%%%%%%%%%%%%%%%%%%% Summary %%%%%%%%%%%%%%%%%%%%%%%%%%%%%%

\textit{NNStreamer} provides pipeline and functional parallelism transparently.
It also allows higher utilization and sharing of different hardware resources virtually without efforts or overheads.
In other words, merely describing the topology with \textit{NNStreamer} enables a higher degree of optimization without system software techniques.
The degree of optimization will be much higher with appropriate system software techniques, time, and efforts; however, such resources are scarce even in large companies.
Results of E4 show the importance of initial design choices as well.

\section{Conclusions}

From the experiences of on-device AI products, we show that the stream processing paradigm may significantly improve performance and productivity.
By deploying \textit{NNStreamer}, we have achieved improvements for on-device AI applications: higher throughput, more straightforward developments with more features and less effort, and improved code quality.
Traditional multimedia developers may also employ arbitrary neural network models in their pipelines with \textit{NNStreamer}.

The lessons learned during the development and deployment of \textit{NNStreamer} include:
\begin{itemize}
    \item Stream processing paradigm works appropriately for on-device AI systems and makes their implementation much more straightforward.
    However, optimizing pipelines still requires some degree of technique and experience.
    Placing and configuring queues, branching and merging, and choosing proper filters for given operations have been sometimes not so trivial.
    \item Showing that a new framework improves performance and efficiency alone is not enough for products or platforms to adopt it.
    As long as conventional implementations meet functional requirements, product engineers are reluctant to adopt a new framework even if costs per application grow excessively.
    We have integrated the work into the software platform along with APIs, user manuals, automated build and deployment systems, test cases, and sample applications.
    We have written first versions of applications and systems, including hardware adaptations with product engineers.
    Then, showing higher productivity, performance, and efficiency has worked.
    \item Analyzing pipeline performance is often complicated and requires specialized tools for visualization and profiling. Training developers with such tools may be required.
    \item For a framework helping applications, developer relations of both public and in-house is crucial.
    Besides, for in-house developers, releasing it as an open-source helps break down silos.
    \item Open sourcing a framework along with opened processes and governance  may appear to incur more workloads.
    However, such workloads include better documentation, rules, policies, broader test cases, and public CI/CD systems, which help improve the overall code quality.
    Besides, we have received more bug and test reports, usage cases, documentation, and code updates.
\end{itemize}

This work is being deployed for various commercial products and platforms and is actively and continuously developed with various future goals.
\textit{NNStreamer} is the standard on-device AI framework of Tizen, which is the OS for a wide range of devices.
We are also deploying \textit{NNStreamer} for Android products.
Developers may install ready-to-use binary packages for Android Studio (JCenter), Ubuntu (PPA), OpenEmbedded (layer), macOS (Homebrew), and Tizen Studio (pre-installed).

%%%%%%%%%%%%%%%%%%%%%% "APPENDIX" FROM HERE %%%%%%%%%%%%%%%%%%%%%%%%%%%%%%%%%%%

%Required by Camera-Ready-Version
%\subsection{Acknowledgements}
%\subsection{Funding Disclosures}
%The CI/CD infrastructure and major contributors are funded by Samsung Research since the start of the project.
%Linux Foundation AI has been supporting communication and documentation infrastructure for the project since April 2020.

\section*{Broader impact}

Initially, we have designed \textit{NNStreamer} for autonomous vehicles.
We have soon discovered that it is applicable for any devices that process neural networks for online data: i.e., analyzing live video streams.
Then, we have successfully deployed \textit{NNStreamer} to Tizen 5.5 (2019.10) and 6.0 (2020.5) as its standard machine learning framework and API.
Tizen is an operating system for general consumer electronics, including mobile phones, TV sets, wearable devices, robotic vacuums, refrigerators, smart ovens, IoT devices, and so on.
For example, the first product using Tizen 5.5, Galaxy Watch 3, uses \textit{NNStreamer} for its on-device AI applications including ``Smart Reply''.
We are applying \textit{NNStreamer} to next-generation Android phones and different consumer electronics in the affiliation; thus, we expect to see its actual usage in large volume immediately, as well.
By making \textit{NNStreamer} compatible with ROS and OpenEmbedded/Yocto, we are providing \textit{NNStreamer} to robotics communities as well.

We have opened the developmental processes and released source codes and binary packages for various platforms as an open-source project.
Therefore, anyone may adopt \textit{NNStreamer} freely for their research or products and contribute to \textit{NNStreamer}.
We already have a few third party companies applying \textit{NNStreamer} for their own products and prototypes.

We have found a significant concern for on-device AI projects.
Machine learning experts are usually not interested in writing optimized or maintainable codes for embedded devices.
Moreover, often, system programmers are also not interested in analyzing and re-implementing codes written by such experts. 
With \textit{NNStreamer}, an easy-to-use pipeline framework for AI projects, we hope to close the gaps between the two parties by adopting pipeline topology as a communication protocol.
Note that this is not necessarily limited to on-device AI projects but also applicable to general server/workstation-based AI projects.

We expect that \textit{NNStreamer} will help improve the productivity of AI researchers in general.
However, the learning curve of describing appropriate pipelines exists, and profiling pipeline performance issues require proper tools and some experiences along with some understandings in queuing theory.
Fortunately, when we have trained developers who have just received their bachelor's degrees in computer science, their learning curves have not been too steep.
They have started writing appropriate pipelines within a few days and optimized pipelines within a couple of weeks.
To further assist novice developers, we are implementing pipeline visualization tools and provide pipeline performance profiling tools.

\textit{NNStreamer} extensions for Protobuf and Flatbuf provide standard representations of tensor streams to interconnect heterogeneous pipelines of MediaPipe and DeepStream.
We can construct pipelines across sensor nodes, edge and mobile devices, workstations, and cloud servers of different stakeholders, which are often referred as ``Edge-AI''.

Stream processing does not need to be restricted to inferences, but can be extended for training.
We are implementing \textit{NNTrainer} (\url{https://github.com/nnstreamer/nntrainer}) so that we can apply on-device training with \textit{NNStreamer} for personalization (adapting a pre-trained neural network for specific users) with personal data kept in the device.
The initial version of \textit{NNTrainer} is already being deployed for personalization services of next-generation products along with \textit{NNStreamer} pipelines.
Besides, stream processing does not need to be restricted to on-device AI application, but can be extended to cloud or server-based AI applications.

There are a few services developed by other affiliations preparing server-based AI services with \textit{NNStreamer} as well as on-device AI systems and devices of different affiliations.
We do not have any contracts or relations with such affiliations except for sharing the same GitHub repository and communicating with them via public channels.
We gather additional requirements from developers of such affiliations and accept contributions from them as well.
In the course of such open collaboration we could have the following benefits:
\begin{itemize}
    \item We have been further required to consider the extensibility, which allowed \textit{NNStreamer} to be adopted to new products in the affiliation that the authors have not imagined.
    \item We have received extensive usage examples and test results from users across various affiliations, which helps improve the functionalities and the robustness.
    \item We have received bug fixes, example applications, new features, and documentations from different affiliations.
    \item Developers have become more enthusiastic with open source software developmental environments. In such environments, daily work including the codes and documents are exposed to the public and the developers are supposed to communicate with developers from different affiliations.
\end{itemize}
We could enjoy such benefits opening not only the source code, but also the whole developmental and policy making processes.
We hope that we could get more contributors and users for the \textit{NNStreamer} project and, someday, we could have voting members and committers in its technical steering committee from various affiliations.
Then, we expect that such higher degree of public inter-affiliation collaboration will help improve both functional and non-functional properties of \textit{NNStreamer} and its sub-projects greatly.

\section*{Availability}

Readers may visit our GitHub pages to get the full source code and its history, sample application code, binary packages, documentations.
We welcome everyone to join \textit{NNStreamer}'s public events, to contribute code commits, to use \textit{NNStreamer} for any purposes, or to discuss via various channels.

\begin{itemize}
\item Web page: \url{https://nnstreamer.ai}
\item GitHub main: \url{https://github.com/nnstreamer/nnstreamer}
\item Gitter: \url{https://gitter.im/nnstreamer}
\item Slack: \url{http://nnstreamer.slack.com}
\item Mailing list: \url{https://lists.lfai.foundation/g/nnstreamer-technical-discuss/}
\\
\item Tizen: Machine Learning APIs are \textit{NNStreamer} APIs.
\item Ubuntu PPA: \texttt{ppa:nnstreamer/ppa}
\item Android Studio: JCenter ``\texttt{nnstreamer}''
\item Yocto/OpenEmbedded Layer: \texttt{meta-neural-network}
\item MacOS Homebrew: \texttt{nnstreamer (\#5926)}
\\
\item Sample: \url{https://github.com/nnstreamer/nnstreamer-example}
\item ROS extension: \url{https://github.com/nnstreamer/nnstreamer-ros}
\end{itemize}

%%%\begin{acks}
%%%We appreciate the issue reports, code submissions, and code reviews of all contributors.
%%% Acknowledgement will be added for CAM-READY version, not in the for-review version.
% ACK, Major Contributors/Idea-Discussion: HongSeok Kim, JinHyuck Park, Sewon Oh
% ACK, Other Contributors: abcinje, niklasjang, mojunsang26, l2zz, mchehab, lemmaa, hs2704sung, tunalee, lee-wonjun, boxerab
% MTCNN-ROS (Sangjung Lee, Mijung Park, ...) & Project:NS Developers (Cho, ...)
% Sponsors (SR managers) (DH Kim, JM Jung, DI Kim, SH Cho)
% Anyone who gave inputs in mailing lists, github issues.
% Communities: GStreamer, Tizen, MediaPipe.
%%%\end{acks}

\section*{Acknowledgements}

This project is funded by Trinity On-Device AI Computing 2021 (RAJ0121ZZ-35RF) of Samsung Research.
We appreciate all the contributors for their issue reports, codes, code reviews, and ideas: HongSeok Kim, JinHyuck Park, Sewon Oh, (@: GitHub ID) @abcinje, @niklasjang, @mojunsang26, @l2zz, @mchehab, @lemmaa, @hs2704sung, @tunalee, @lee-wonjun, @boxerab, and many others.
We appreciate the support of Linux Foundation AI \& Data.
Authors would like to thank Daehyun Kim, Jinmin Chung, Duil Kim, JeongHoon Park, Seunghwan Cho, and Sebastian Seung of Samsung Research for the encouragements and fruitful discussions.

{\normalsize \bibliographystyle{acm}
\bibliography{nnstreamer}}

\end{document}